\documentclass[a4paper,twoside]{article}

\usepackage{epsfig}
\usepackage{subcaption}
\usepackage{calc}
\usepackage{amssymb}
\usepackage{amstext}
\usepackage{amsmath}
\usepackage{amsthm}
\usepackage{multicol}
\usepackage{pslatex}
\usepackage{apalike}
\usepackage{algorithm2e}
\usepackage[bottom]{footmisc}
\usepackage[inline]{enumitem}

\usepackage[textsize=tiny]{todonotes}
\usepackage{graphicx}
\usepackage[caption=false,font=footnotesize]{subfig}
\usepackage{multirow}
\usepackage{booktabs,tabularx}
\usepackage{siunitx}
\usepackage{svg}

\newcommand{\thickhline}{\noalign{\hrule height 1.5pt}}

\usepackage{SCITEPRESS}     

\begin{document}

\title{RNM-TD3: N:M Semi-structured Sparse Reinforcement Learning From Scratch}
\author{\authorname{Isam Vrce \sup{1}\orcidAuthor{0009-0000-1722-2812}, Andreas Kassler\sup{1}\orcidAuthor{0000-0002-9446-8143} and Gökçe Aydos\sup{2}\orcidAuthor{0000-0002-4183-9086}}
\affiliation{\sup{1}Deggendorf Institute of Technology, Dieter-Görlitz-Platz 1, Deggendorf, Germany}
\affiliation{\sup{2}Department of Engineering Technology, Technical University of Denmark, Ballerup, Denmark}
\email{\{isam.vrce, andreas.kassler\}@th-deg.de, gokay@dtu.dk}
}


\abstract{
Sparsity is a well-studied technique for compressing deep neural networks (DNNs) without compromising performance. In deep reinforcement learning (DRL), neural networks with up to 5\% of their original weights can still be trained with minimal performance loss compared to their dense counterparts. However, most existing methods rely on unstructured fine-grained sparsity, which limits hardware acceleration opportunities due to irregular computation patterns. Structured coarse-grained sparsity enables hardware acceleration, yet typically degrades performance and increases pruning complexity. In this work, we present, to the best of our knowledge, the first study on $N{:}M$ structured sparsity in RL, which balances compression, performance, and hardware efficiency. Our framework enforces row-wise $N{:}M$ sparsity throughout training for all networks in off-policy RL (TD3), maintaining compatibility with accelerators that support $N{:}M$ sparse matrix operations. Experiments on continuous-control benchmarks show that RNM-TD3, our $N{:}M$ sparse agent, outperforms its dense counterpart at 50\%-75\% sparsity (e.g., 2:4 and 1:4), achieving up to a 14\% increase in performance at 2:4 sparsity on the Ant environment. RNM-TD3 remains competitive even at 87.5\% sparsity (1:8), while enabling potential training speedups.
}

\onecolumn \maketitle \normalsize \setcounter{footnote}{0} \vfill

\section{\uppercase{Introduction}}
\label{sec:introduction}

\begin{figure*}[ht]
\vskip 0.2in
\begin{center}
\centerline{\includegraphics[width=\linewidth]{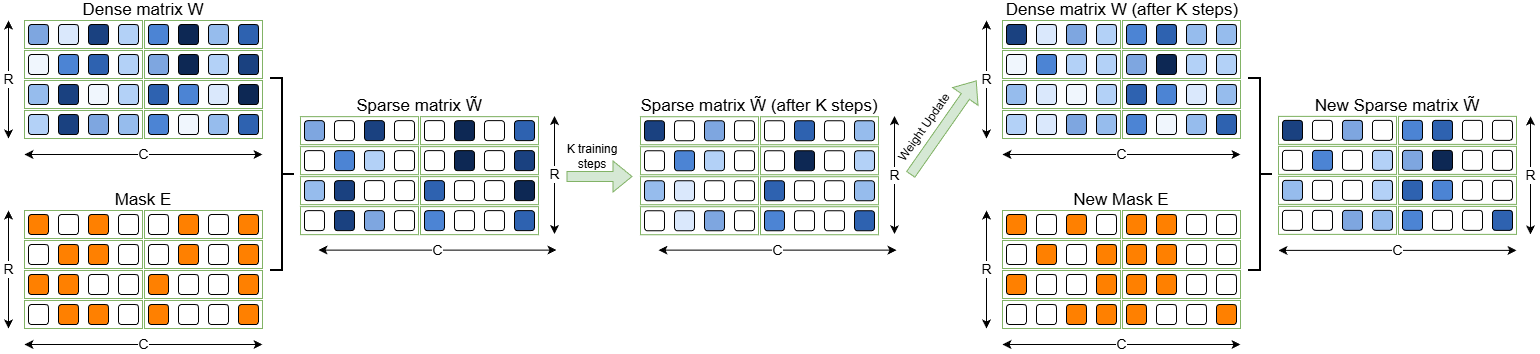}}
\caption{Illustration of a single cycle of the RNM-TD3 algorithm, demonstrated with an $N:M$ sparsity pattern where $N=2$ and $M=4$. The weight matrices depicted have dimensions $R \times C$ (rows $\times$ columns). (Left) A dense matrix $W$ and an $N:M$ sparse mask $E$ are combined to create the sparse matrix $\tilde{W}$. This sparse matrix is trained for $K$ steps, during which only the active weights update the underlying dense matrix. (Right) After $K$ training steps, the weights in the original matrix $W$ change in magnitude (only at the positions of active weights), resulting in a different mask $E$. The updated matrix $W$ and the new mask $E$ create a new sparse matrix $\tilde{W}$. This cycle repeats until the end of training. 
}
\label{fig:explanation}
\end{center}
\vskip -0.2in
\end{figure*}

Modern deep learning (DL) largely relies on over-parameterized networks, whose performance scales with data, model size, and compute \cite{hestness2017deep,kaplan2020scaling}. Yet, practical deployment demands low latency, low energy consumption, and small memory footprints. Model compression techniques, such as pruning/quantization \cite{han2015deep,gale2019state}, distillation \cite{hinton2015distilling}, and sparse training \cite{frankle2018lottery,evci2020rigging}, address these challenges by reducing model size while preserving accuracy. In this work, we focus on sparse training in DRL.

A neural network is considered sparse when a large fraction of its parameters are zero. Based on the distribution of zero-valued parameters, sparsity can be classified into three categories: \textbf{Unstructured sparsity} zeroes individual weights distributed across the network without any structure, often achieving high compression with minimal accuracy loss \cite{zhu2017prune,molchanov2017variational,frantar2023sparsegpt}. However, the irregular pattern limits hardware speedups. \textbf{Structured sparsity} removes entire channels, filters, blocks. This process shrinks tensors, which reduces the number of multiply and accumulate (MAC) operations. However, coarser granularity often yields lower accuracy than unstructured sparse networks at the same sparsity level \cite{cheng2024survey}, with some mitigation via dependency-aware pruning at an additional computational cost \cite{fang2023depgraph}. \textbf{Semi-structured $N{:}M$ sparsity} enforces a constraint whereby each contiguous group of $M$ weights contains no more than $N$ nonzero weights \cite{zhou2021learning}. When supported in hardware (e.g., 2:4 on NVIDIA Ampere Tensor Cores), this regularity enables accelerated matrix multiplications during the forward pass \cite{nvidia2021nm-sparsity}. However, accelerating backpropagation requires an $N{:}M$ sparsity pattern along the columns of the original weight matrix $W$ \cite{zhang2023bi,Chenyi2025BAME}. Beyond the 2:4 sparsity pattern, finer sparsity patterns such as 1:3 and 1:4 have also been explored \cite{taka2025systolic}, enabling greater flexibility.

In contrast to supervised learning, DRL involves training on non-stationary data induced by a continually evolving policy, using bootstrapping with delayed rewards and moving targets \cite{sutton2018reinforcement,mnih2015human}. These characteristics make DRL training inherently unstable, and pruning can further exacerbate this instability \cite{livne2020pops}. Existing sparse DRL approaches either prune pre-trained dense models \cite{yu2019playing} or employ topology-evolution strategies that involve neuron pruning and regrowth \cite{sokar2021dynamic,tan2022rlx2}. While effective for compression, both strategies limit training-time acceleration, as post-training pruning does not exploit sparse matrices during training.

In this work, we introduce an end-to-end $N{:}M$ structured sparse DRL framework that maintains a hardware-aware $N{:}M$ sparsity pattern throughout training for both actor and critic networks of the state-of-the-art DRL algorithm Twin Delayed Deep Deterministic Policy Gradient (TD3). We demonstrate that the period of mask updates correlates with Sparse Architecture Divergence (SAD) \cite{zhou2021learning}, providing new insights into the dynamics of sparsity during DRL training. Results indicate that training an $N{:}M$ sparse agent can match the performance of dense and unstructured baselines. We make the following contributions:
\begin{itemize}
    \item \textbf{Framework for $N{:}M$-sparse DRL training.} We maintain strict $N{:}M$ constraints throughout training so that both actor and critic are compatible with hardware-aware sparse matrix-matrix multiplications. (e.g., 2:4 on NVIDIA Ampere Tensor Cores) \cite{nvidia2021nm-sparsity}.
    \item \textbf{SAD mask dynamics study.} We relate the period of mask update to the SAD between successive masks. We find that DRL favors less frequent mask updates compared to supervised learning settings. Moreover, we show that maintaining a stable, non-zero SAD, which is achieved naturally through training, is correlated with improved cumulative reward.
    \item \textbf{Empirical validation.} We demonstrate that on MuJoCo continuous-control benchmarks, our RNM-TD3 agent outperforms dense baselines on average return and rivals unstructured sparse baselines, while offering theoretical acceleration on hardware that supports $N{:}M$ sparse computation.
\end{itemize}

The rest of the paper is structured as follows. Section 2 reviews related work. Section 3 details our methodology and introduces our sparse training strategy. We evaluate our approach in Section 4 and conclude the paper in Section 5.

\section{\uppercase{Related Work}}
Early work on compressing neural networks in RL focused on knowledge distillation. A trained agent served as the source model, and its behavior was distilled into a compact student network, resulting in a model size reduction of up to 15 times with no performance loss \cite{rusu2015policy}. Building on this, \cite{schmitt2018kickstarting} introduced an unconstrained teacher-student paradigm, reporting a 42\% performance gain while requiring $\sim 10\times$ fewer training steps for the student.

The first attempts to induce sparsity in DRL focused on post-training pruning: (i) train a dense model, (ii) remove small-magnitude weights, (iii) fine-tune the sparse model. This strategy closely mirrors iterative magnitude pruning techniques developed in deep supervised learning \cite{han2015learning}, and was later adapted to DRL \cite{livne2020pops}.

Works inspired by the lottery ticket hypothesis (LTH) \cite{frankle2018lottery} applied in RL show that sparse sub-networks, when trained from scratch, can match or even surpass their dense counterparts \cite{yu2019playing,vischer2021lottery}. In offline RL, single-shot pruning methods (i.e., sparsity is defined at initialization and remains fixed throughout training) such as SNIP \cite{lee2018snip} and GraSP \cite{wang2020picking} remain competitive even at $\sim 95\%$ sparsity \cite{arnob2021single}. More recently, \cite{ma2025network} adopted a pre-training pruning scheme based on one-shot random pruning, showing that highly sparse DRL networks can outperform dense baselines.

Dynamic Sparse Training (DST) trains sparse models from scratch by dynamically adapting the network topology, pruning low-importance weights and regrowing new connections during training. While static sparse training can be surprisingly effective in RL \cite{graesser2022state}, DST methods \cite{mocanu2018scalable,evci2020rigging} aim to optimize performance by evolving the structure alongside the weights. RLx2 \cite{tan2022rlx2} achieves a compression of $7.5\times$ to $20\times$ with $<3\%$ performance loss compared to the original dense counterpart.  

Following a similar philosophy of dynamic topology adaptation, \cite{sokar2021dynamic} learn binary masks to select sparse sub-networks, optimizing the topology via periodic updates. Extending the concept of learning sub-networks within a dense network, \cite{arnob2024efficient} discover binary task-specific masks, enabling multiple specialized pathways within a single model.

Most existing sparse DRL approaches focus primarily on two objectives: model compression and optimizing policy performance under sparsity constraints. However, these methods typically rely on unstructured sparsity, which suffers from irregular memory access patterns and fails to translate theoretical reductions in floating-point operations (FLOPs) into real-world speedups on standard hardware \cite{wen2016learning}. In contrast, we aim to optimize on three fronts simultaneously: model compression, policy quality, and training acceleration. Our method enforces hardware-aware $N{:}M$ sparsity from the beginning of the training with no dense initialization. This ensures that the computational efficiency of the $N{:}M$ pattern is utilized throughout the entire training, enabling acceleration from the very first step.

\section{\uppercase{Methodology}}
In this section, we present RNM-TD3, an algorithm designed to train a row-wise $N{:}M$ structured sparse agent from scratch. We build on TD3, a widely used off-policy actor–critic algorithm. Using the deterministic TD3 algorithm as our backbone enables us to isolate the impact of $N{:}M$ structured sparsity on performance without the additional factors introduced by stochastic policy methods. Figure \ref{fig:explanation} provides a high-level description of the key idea, how the row-wise $N{:}M$ structured sparsity is maintained during training. We first introduce the problem formulation and notation, describe our approach in more detail, highlight the key differences between sparse training in DRL and supervised learning and our method to stabilize training through soft-reset.

\subsection{Problem Formulation \& Notation}
We frame the RL problem as a standard Markov decision process (MDP) $\mathcal{M} = (\mathcal{S}, \mathcal{A}, \mathcal{P}, r, \gamma)$. Here, $\mathcal{S}$ denotes the state space, $\mathcal{A}$ is the continuous action space, $\mathcal{P}$ represents the transition probability distribution, $r$ is the reward function, and $\gamma \in [0,1)$ is the discount factor.

The TD3 agent consists of a deterministic actor $\mu_\theta(s)$ and twin critics $Q_{\phi_1}, Q_{\phi_2}$, parameterized by $\theta$ and $\phi_1, \phi_2$, respectively. To enable hardware acceleration, we enforce a row-wise $N{:}M$ sparsity pattern, meaning that for each row of a weight matrix and each group of $M$ consecutive weights in that row, at most $N$ weights are non-zero.

Let $W^{(\ell)} \in \mathbb{R}^{d_{\text{out}} \times d_{\text{in}}}$ denote the weight matrix of a fully connected layer $\ell$, where $d_{\text{in}}$ and $d_{\text{out}}$ represent the number of input and output neurons, respectively. Our objective is to maximize the expected return while ensuring that $W^{(\ell)}$ satisfies the row-wise $N{:}M$ sparsity pattern:

\begin{equation}
\begin{split}
\forall\, i\in\{1,\dots,d_{\text{out}}\},\;
\forall\, j\in\{1,\dots,\lceil d_{\text{in}}/M\rceil\}:\quad \\
\bigl\|W^{(\ell)}_{\,i,\,(j-1)M+1:\,\min\{jM,\,d_{\text{in}}\}}\bigr\|_0 \le N .
\end{split}
\label{eq:rowwise-nm-constraint}
\end{equation}

\noindent where $\|\cdot\|_0$ denotes the $L_0$ norm, which counts the number of non-zero elements. 

Intuitively, Eq.~\eqref{eq:rowwise-nm-constraint} partitions each row of \(W^{(\ell)}\) into contiguous blocks of size $M$. Within each block, at most $N$ entries are non-zero. Sparsity is enforced via binary masks \(E^{(\ell)}\in\{0,1\}^{d_{\text{out}}\times d_{\text{in}}}\), yielding effective weights \(\widetilde{W}^{(\ell)} = W^{(\ell)} \odot E^{(\ell)}\). The projection operator \(\mathcal{P}_{N{:}M}\) used to construct the mask is defined as:

\begin{equation}
\mathcal{P}_{N{:}M}(W)_{i,j} =
\begin{cases}
1, &
\text{if }
\begin{array}[t]{@{}l@{}}
|W_{i,j}| \text{ is in the top-}N \\
\text{ of block } (i,\lfloor (j-1)/M \rfloor)
\end{array}
\\[6pt]
0, & \text{otherwise.}
\end{cases}
\label{eq:projection}
\end{equation}

Masks are initialized to satisfy Eq.~\eqref{eq:rowwise-nm-constraint} and are applied to all six TD3 networks (i.e., the online actor network, the twin critic networks, and their corresponding target networks). Online masks are updated periodically during training, while remaining fixed between updates. Target networks are updated by Polyak averaging \cite{lillicrap2015continuous}:

\begin{equation}
\bar\psi \leftarrow \tau\,\psi + (1-\tau)\,\bar\psi,
\qquad \psi \in \{\theta,\phi_1,\phi_2\}.
\label{eq:soft-update}
\end{equation}
\noindent
where $\tau$ is a small constant (typically $\tau= 0.005$) that controls the interpolation rate. However, during the online mask update, the online mask is immediately copied to the corresponding targets to maintain stability.

\subsection{Training under $N{:}M$ Sparsity}
Training starts with $N{:}M$ sparse neural networks. The weight matrix $W$ is initialized using a Kaiming uniform distribution, adjusted to account for the reduced effective fan-in induced by the $N{:}M$ sparsity constraint. Immediately after initialization and prior to the first training step, we apply the projection operator from Eq.~\eqref{eq:projection}:
\begin{equation}
E = \mathcal{P}_{N:M}(W), \qquad \widetilde{W} = W \odot E.
\end{equation}

The resulting sparse matrix $\widetilde{W}$ is now N:M sparse. To update the weights in the dense weights $W$, we propagate the gradients from the loss $\mathcal{L}$ only to the active weights, while the gradients for inactive weights are zeroed out:

\begin{equation}
\frac{\partial \mathcal{L}}{\partial W}
=
\frac{\partial \mathcal{L}}{\partial \widetilde{W}} \odot E.
\label{eq:masked-grad}
\end{equation}

After K training steps, if the magnitude of an active weight becomes smaller than that of any inactive weight within the same group of $M$ weights, the inactive weight with the largest magnitude is reactivated at the next mask update, thereby altering the network topology.

However, this approximation can induce unstable mask updates, often referred to as "mask flickering". To quantify topological evolution, we measure the Sparse Architecture Divergence (SAD) \cite{zhou2021learning}, defined as the Hamming distance between consecutive masks:
\begin{equation}
    \mathrm{SAD}_{t, t+k} = \| E_{t+k} - E_t \|_{1}.
\end{equation}
High SAD indicates significant topological changes, while low SAD suggests convergence to a fixed sparsity pattern. While supervised learning typically benefits from minimizing SAD to ensure convergence \cite{zhou2021learning,Chenyi2025BAME}, our analysis suggests that DRL requires a different regime due to the non-stationary data distribution.

\subsection{Differences Between $N{:}M$ Sparse Training in DRL and Supervised Learning}

In supervised learning, $N{:}M$ sparse training typically updates masks at every optimization step. To reduce SAD between successive mask updates and improve final accuracy, supervised learning techniques introduced a weight factor $\lambda_W$ that progressively drives inactive weights toward zero at each step: Sparse Refined STE (SR-STE) \cite{zhou2021learning}.

\begin{figure}[!h]
  \centering
   {\epsfig{file = 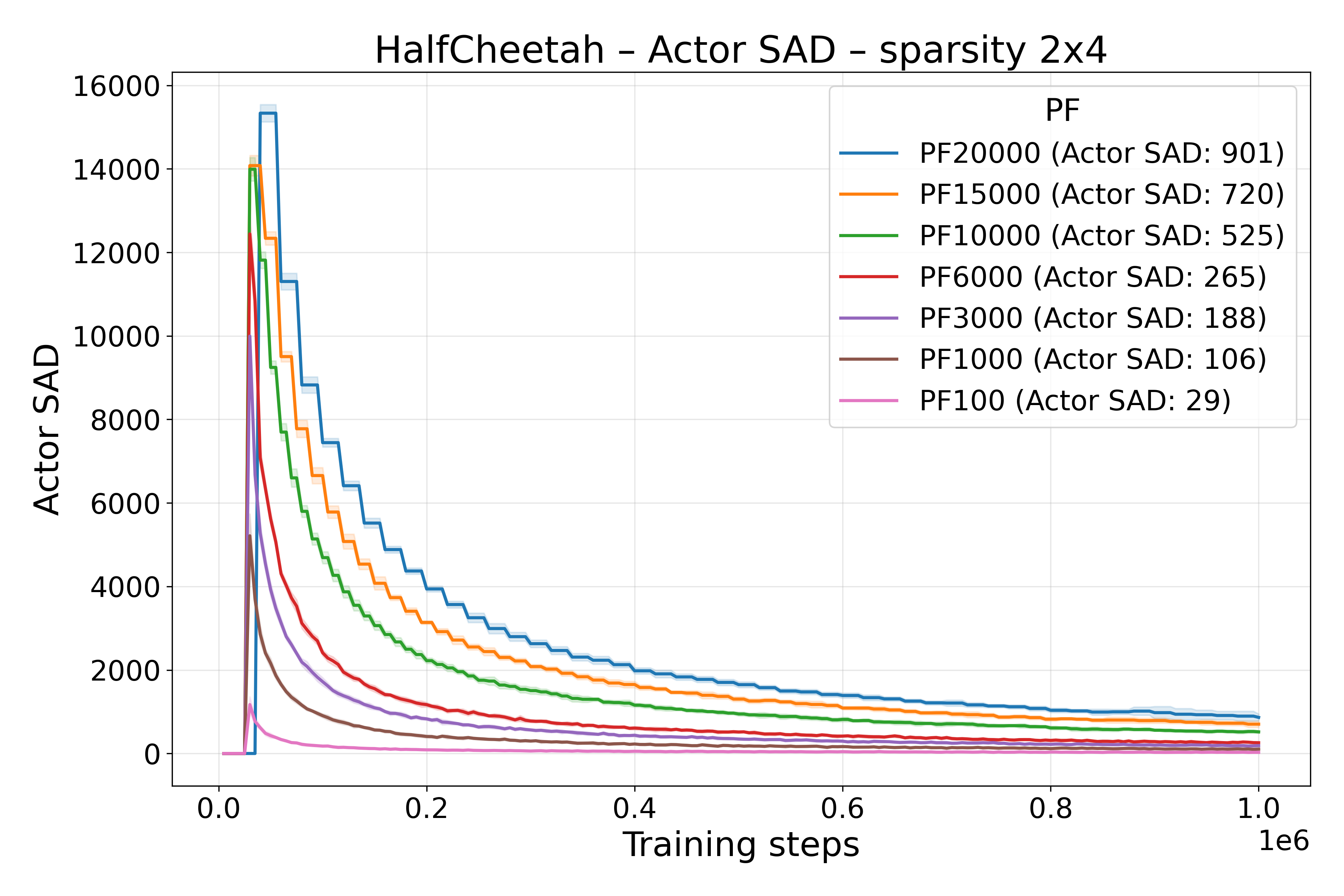, width=\columnwidth}}
  \caption{SAD for 7 different mask update periods for the actor network in the Half Cheetah environment. SAD decays approximately exponentially regardless of the mask update period.}
  \label{fig:example1}
 \end{figure}

In DRL, however, we find that superior performance is achieved with much less frequent mask updates, occurring every $K \gg 1$ environment steps. A controlled ablation study over $K$ (Section \ref{SAD}) reveals an optimal mask update period of around $K = 4000$ for HalfCheetah environment with 1:4 sparsity. The optimal "K" depends on both the environment and the sparsity level. Values around $K = 5000$ environment steps work well across tasks and sparsity patterns. SAD decays approximately exponentially toward a stable non-zero value (see Figure \ref{fig:example1}), in contrast to the concave decay trajectories observed in supervised learning (see Figure 4a in \cite{zhou2021learning}). This suggests that explicit SAD reduction mechanisms, such as those in SR-STE, are unnecessary or even detrimental in the DRL setting. We examined two natural variants of applying the SR-STE shrinkage factor $\lambda_W$ within DRL:

\begin{enumerate}
    \item \textbf{Per-step application.}
    Applying $\lambda_W$ at every training step causes inactive weights that are masked by chance at initialization to be repeatedly shrunk for the next $K$ steps (i.e., until the next mask update). This acts as an one-shot pruning of potentially important connections, leading to the premature suppression of useful weights and degraded performance.

    \item \textbf{Per-mask-update application.}
    Applying $\lambda_W$ only once when the mask is updated avoids per-step shrinkage. However, in momentum-based optimizers (e.g., Adam \cite{Kingma2014AdamAM}), this factor still contributes across many subsequent updates. Since the gradient for inactive weights is zero, $\lambda_W$ becomes the only factor affecting their magnitude, progressively driving them toward zero between mask updates. This reduces the likelihood of weight reactivation and, consequently, limits topological flexibility.
\end{enumerate}

Our empirical results suggest that in DRL, the goal is not to drive SAD to zero. Instead, a stable, non-zero SAD emerges naturally when masks are updated at the appropriate frequency. Configurations with larger SAD exhibit greater adaptability to changing policy dynamics, whereas frequent mask updates result in low SAD, which reduces topological flexibility and harms final performance.

\subsection{Stabilizing $N{:}M$ Training Through Soft Reset}
Training highly sparse $N{:}M$ networks in inherently unstable environments, such as \textit{Humanoid}, can lead to training instabilities. The combination of an unstable environment, high sparsity ratios (e.g., 1:8), and dynamic sparse training from initialization often results in a network topology that lacks the capacity to model complex locomotion dynamics during the early stages of training. There is a high probability of severing the information flow from key sensory inputs (e.g., knee angle sensors) to critical actuators (e.g., leg motors) by masking important weights. Consequently, the agent attempts to maintain balance by activating non-critical actuators, such as those in the arms, which fails to produce meaningful locomotion. The agent falls shortly after the start of the episode, generating a series of short episodes that prevent the accumulation of meaningful learning experiences.

After a mask change, the critic suggests target values for the actor, which the actor attempts to match by minimizing the error. However, due to high sparsity, the few remaining active connections must compensate for the missing connections. This is especially problematic at the beginning of training when weights are randomly initialized. To achieve the output shifts required by the critic using such limited connectivity, the actor aggressively increases the magnitude of the active weights. Given the $\tanh$ nonlinearity at the output, these high-magnitude pre-activations cause the neurons to saturate (locking at $\pm 1$). As a result, the agent repeatedly selects identical actions, reducing the SAD to zero.

Empirically, with a mask-update period of $K=15{,}000$, early performance collapse occurred in approximately 80\% of runs in the \textit{Humanoid} task. To address this, we introduce a simple \emph{soft reset} mechanism. Whenever SAD falls below a critical threshold (e.g., $\mathrm{SAD} < 10$) during a mask update, we resample a new random $N{:}M$ mask. This forces topological exploration. The soft reset completely eliminates early collapse. After recovery, training proceeds normally, and SAD maintains a healthy non-zero value throughout the remainder of the training.

\section{\uppercase{Experimental Evaluation}}

\begin{table*}[t]
\caption{Comparison of RNM-TD3 with baselines. Results are reported as mean $\pm$ standard deviation, normalized relative to the dense baseline (i.e., scaled such that the dense mean and standard deviation are 1.0). Raw dense baseline scores are provided for reference: Ant-v5 ($4453\pm1320$), HalfCheetah-v5 ($10211\pm1110$), Humanoid-v5 ($4868\pm659$), and Walker2d-v5 ($3767\pm833$). TS denotes Target Sparsity ($\%$). RNM-TD3 and SNM has the same realized sparsity which slightly deviates from TS because the last layer remains dense (note that last layer is dense in all methods). DS-TD3 method uses the same layer sparsity parameter for actor and critic networks, so realized sparsity will always differ from target sparsity.}
\label{tab:example2}
\centering
\renewcommand{\arraystretch}{1}
\begin{tabular}{|c|c|cc|cc|cc|c|}
\hline
\multirow{2}{*}{Environment} & \multirow{2}{*}{TS(\%)} & \multicolumn{2}{c|}{RNM-TD3(best)}                                     & \multicolumn{2}{c|}{SSN-N:M (best)}                                      & \multicolumn{2}{c|}{DS-TD3}                     & RLx2                    \\
                             &                         & Act.(\%)                                & Crit.(\%)               & Act.(\%)                                 & Crit.(\%)               & Act.(\%)                     & Crit.(\%)   & Act./Crit.(\%)        \\ \thickhline
Ant-v5                       & \multirow{2}{*}{50}     & \multicolumn{2}{c|}{\(\,1.14\,\pm\,0.05\,{\scriptstyle (2{:}4)}\,\)} & \multicolumn{2}{c|}{\(\,1.05\,\pm\,0.21\,{\scriptstyle (4{:}8)}\,\)} & \multicolumn{2}{c|}{\(\,0.91\,\pm\,0.18\,\)} & \(\,0.93\,\pm\,0.16\,\) \\ \cline{3-9} 
                             &                         & \multicolumn{1}{c|}{48.92}               & 49.86                    & \multicolumn{1}{c|}{48.92}                & 49.86                    & \multicolumn{1}{c|}{48.24}    & 49.94        & 50                      \\ \cline{2-9} 
                             & \multirow{2}{*}{75}     & \multicolumn{2}{c|}{\(\,1.07\,\pm\,0.15\,{\scriptstyle (1{:}4)}\,\)} & \multicolumn{2}{c|}{\(\,0.93\,\pm\,0.28\,{\scriptstyle (1{:}4)}\,\)} & \multicolumn{2}{c|}{\(\,0.78\,\pm\,0.16\,\)} & \(\,1.01\,\pm\,0.12\,\) \\ \cline{3-9} 
                             &                         & \multicolumn{1}{c|}{73.37}               & 74.79                    & \multicolumn{1}{c|}{73.37}                & 74.79                    & \multicolumn{1}{c|}{72.84}    & 74.64        & 75                      \\ \cline{2-9} 
                             & \multirow{2}{*}{87.5}   & \multicolumn{2}{c|}{\(\,0.86\,\pm\,0.16\,{\scriptstyle (1{:}8)}\,\)} & \multicolumn{2}{c|}{\(\,0.75\,\pm\,0.18\,{\scriptstyle (1{:}8)}\,\)} & \multicolumn{2}{c|}{\(\,0.83\,\pm\,0.19\,\)} & \(\,0.82\,\pm\,0.19\,\) \\ \cline{3-9} 
                             &                         & \multicolumn{1}{c|}{85.60}               & 87.26                    & \multicolumn{1}{c|}{85.60}                & 87.26                    & \multicolumn{1}{c|}{85.34}    & 87.19        & 87.5                    \\ \thickhline
H.Cheetah-v5                 & \multirow{2}{*}{50}     & \multicolumn{2}{c|}{\(\,1.01\,\pm\,0.08\,{\scriptstyle (4{:}8)}\,\)} & \multicolumn{2}{c|}{\(\,0.96\,\pm\,0.08\,{\scriptstyle (2{:}4)}\,\)} & \multicolumn{2}{c|}{\(\,1.05\,\pm\,0.09\,\)} & \(\,0.95\,\pm\,0.10\,\) \\ \cline{3-9} 
                             &                         & \multicolumn{1}{c|}{48.92}               & 49.82                    & \multicolumn{1}{c|}{48.92}                & 49.82                    & \multicolumn{1}{c|}{47.77}    & 49.65        & 50                      \\ \cline{2-9} 
                             & \multirow{2}{*}{75}     & \multicolumn{2}{c|}{\(\,0.90\,\pm\,0.05\,{\scriptstyle (1{:}4)}\,\)} & \multicolumn{2}{c|}{\(\,0.82\,\pm\,0.07\,{\scriptstyle (1{:}4)}\,\)} & \multicolumn{2}{c|}{\(\,0.95\,\pm\,0.17\,\)} & \(\,0.92\,\pm\,0.10\,\) \\ \cline{3-9} 
                             &                         & \multicolumn{1}{c|}{73.39}               & 74.73                    & \multicolumn{1}{c|}{73.39}                & 74.73                    & \multicolumn{1}{c|}{73.00}    & 74.84        & 75                      \\ \cline{2-9} 
                             & \multirow{2}{*}{87.5}   & \multicolumn{2}{c|}{\(\,0.72\,\pm\,0.06\,{\scriptstyle (1{:}8)}\,\)} & \multicolumn{2}{c|}{\(\,0.68\,\pm\,0.05\,{\scriptstyle (1{:}8)}\,\)} & \multicolumn{2}{c|}{\(\,0.81\,\pm\,0.08\,\)} & \(\,0.80\,\pm\,0.05\,\) \\ \cline{3-9} 
                             &                         & \multicolumn{1}{c|}{85.62}               & 87.18                    & \multicolumn{1}{c|}{85.62}                & 87.18                    & \multicolumn{1}{c|}{85.23}    & 87.05        & 87.5                    \\ \thickhline
Humanoid-v5                  & \multirow{2}{*}{50}     & \multicolumn{2}{c|}{\(\,1.05\,\pm\,0.03\,{\scriptstyle (2{:}4)}\,\)} & \multicolumn{2}{c|}{\(\,1.03\,\pm\,0.05\,{\scriptstyle (2{:}4)}\,\)} & \multicolumn{2}{c|}{\(\,1.04\,\pm\,0.02\,\)} & \(\,0.99\,\pm\,0.03\,\) \\ \cline{3-9} 
                             &                         & \multicolumn{1}{c|}{48.63}               & 49.92                    & \multicolumn{1}{c|}{48.63}                & 49.92                    & \multicolumn{1}{c|}{48.16}    & 50.01        & 50                      \\ \cline{2-9} 
                             & \multirow{2}{*}{75}     & \multicolumn{2}{c|}{\(\,1.04\,\pm\,0.03\,{\scriptstyle (1{:}4)}\,\)} & \multicolumn{2}{c|}{\(\,1.01\,\pm\,0.12\,{\scriptstyle (1{:}4)}\,\)} & \multicolumn{2}{c|}{\(\,0.99\,\pm\,0.06\,\)} & \(\,0.93\,\pm\,0.07\,\) \\ \cline{3-9} 
                             &                         & \multicolumn{1}{c|}{72.95}               & 74.88                    & \multicolumn{1}{c|}{72.95}                & 74.88                    & \multicolumn{1}{c|}{72.52}    & 74.73        & 75                      \\ \cline{2-9} 
                             & \multirow{2}{*}{87.5}   & \multicolumn{2}{c|}{\(\,0.28\,\pm\,0.20\,{\scriptstyle (1{:}8)}\,\) *} & \multicolumn{2}{c|}{\(\,0.82\,\pm\,0.35\,{\scriptstyle (1{:}8)}\,\)} & \multicolumn{2}{c|}{\(\,1.02\,\pm\,0.03\,\)} & \(\,0.64\,\pm\,0.26\,\) \\ \cline{3-9} 
                             &                         & \multicolumn{1}{c|}{85.10}               & 87.36                    & \multicolumn{1}{c|}{85.10}                & 87.36                    & \multicolumn{1}{c|}{84.89}    & 87.28        & 87.5                    \\ \thickhline
Walker2d-v5  & \multirow{2}{*}{50}     & \multicolumn{2}{c|}{\(\,1.04\,\pm\,0.13\,{\scriptstyle (4{:}8)}\,\)}   & \multicolumn{2}{c|}{\(\,0.94\,\pm\,0.22\,{\scriptstyle (4{:}8)}\,\)}  & \multicolumn{2}{c|}{\(\,1.18\,\pm\,0.13\,\)} & \(\,0.97\,\pm\,0.15\,\) \\ \cline{3-9} 
                                        &                         & \multicolumn{1}{c|}{48.92}                 & 49.82                     & \multicolumn{1}{c|}{48.92}                 & 49.82                    & \multicolumn{1}{c|}{47.77}    & 49.65        & 50                      \\ \cline{2-9} 
                                        & \multirow{2}{*}{75}     & \multicolumn{2}{c|}{\(\,0.96\,\pm\,0.09\,{\scriptstyle (1{:}4)}\,\)}   & \multicolumn{2}{c|}{\(\,0.89\,\pm\,0.12\,{\scriptstyle (1{:}4)}\,\)}  & \multicolumn{2}{c|}{\(\,0.99\,\pm\,0.26\,\)} & \(\,0.82\,\pm\,0.24\,\) \\ \cline{3-9} 
                                        &                         & \multicolumn{1}{c|}{73.39}                 & 74.73                     & \multicolumn{1}{c|}{73.39}                 & 74.73                    & \multicolumn{1}{c|}{73.00}    & 74.84        & 75                      \\ \cline{2-9} 
                                        & \multirow{2}{*}{87.5}   & \multicolumn{2}{c|}{\(\,0.58\,\pm\,0.29\,{\scriptstyle (1{:}8)}\,\)}   & \multicolumn{2}{c|}{\(\,0.53\,\pm\,0.26\,{\scriptstyle (2{:}16)}\,\)} & \multicolumn{2}{c|}{\(\,0.86\,\pm\,0.30\,\)} & \(\,0.83\,\pm\,0.20\,\) \\ \cline{3-9} 
                                        &                         & \multicolumn{1}{c|}{85.62}                 & 87.18                     & \multicolumn{1}{c|}{85.62}                 & 87.18                    & \multicolumn{1}{c|}{85.23}    & 87.05        & 87.5                    \\ \hline
\end{tabular}
\end{table*}

We evaluate our proposed sparse TD3 variant, \textbf{RNM-TD3}, on continuous-control tasks from the Gymnasium \cite{towers2024gymnasium} MuJoCo benchmark suite \cite{mujoco}. RNM-TD3 introduces a general $N{:}M$ structured sparsity framework built on top of the TD3 algorithm, agnostic to both network architecture and specific hyperparameters. Our experimental study addresses two key questions: (i) how different $N{:}M$ sparsity patterns impact performance across diverse environments relative to dense and unstructured baselines, and (ii) how the mask-update period influences both the return and the dynamics of the sparse topology.

We compare our method against four baselines: 
\begin{itemize}
\item \textbf{Dense}, TD3 with dense neural networks.
\item \textbf{SSN N:M}, a static sparse network with a fixed $N{:}M$ mask sampled at initialization and never updated.
\item \textbf{DS-TD3} \cite{sokar2021dynamic}: A dynamic sparse training method utilizing random regrowth.
\item\textbf{RLx2} \cite{tan2022rlx2}: A method that prunes weights with the smallest magnitude and regrows connections based on gradient magnitude.
\end{itemize}

To ensure a fair comparison, all methods use identical architectures and hyperparameters. The actor and both critic networks consist of two hidden layers with 256 neurons. All experiments run for 1M environment steps using a standard static replay buffer and 1-step returns. Except for their specific sparsity mechanisms, all algorithms share identical TD3 settings (e.g., learning rate, batch size). Performance is reported as the average return over 11 random seeds for each environment over 1M steps.

\subsection{Comparative evaluation on continuous RL tasks}

\begin{figure*}[ht]
\vskip 0.2in
\begin{center}
\centerline{\includegraphics[width=\linewidth]{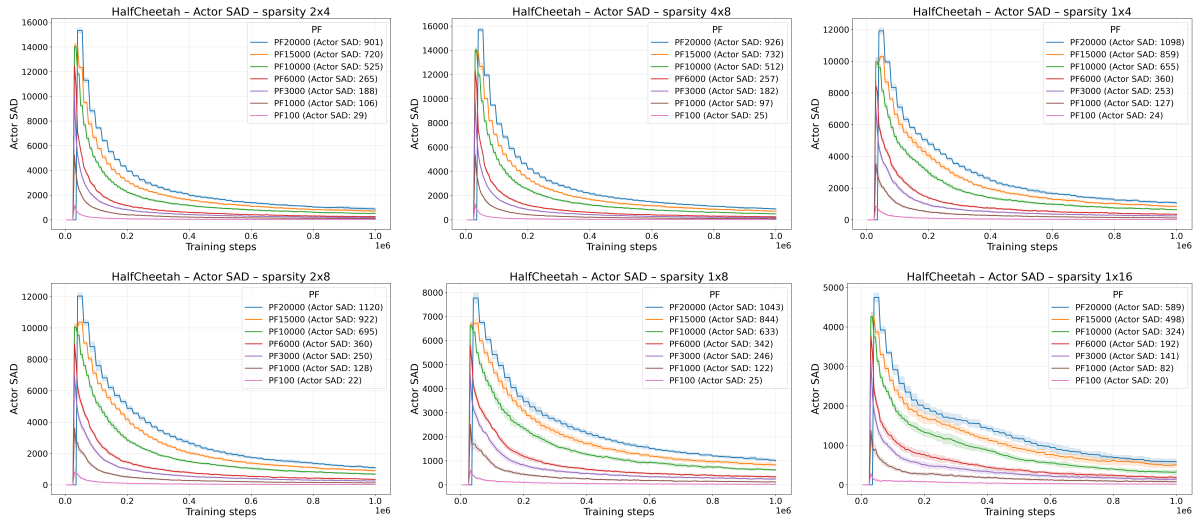}}
\caption{SAD for the HalfCheetah actor network for different sparsity patterns and mask update periods. Each curve is averaged over 11 seeds. Overall 462 runs (7 update periods $\times$ 6 sparsity patterns $\times$ 11 seeds) confirm a clear correlation between the mask update period and SAD: more frequent mask updates (smaller update periods) lead to lower SAD, while less frequent updates (larger update periods) result in higher SAD.}
\label{fig:HC Actor SAD}
\end{center}
\vskip -0.2in
\end{figure*}

As shown in Table \ref{tab:example2}, RNM-TD3 exhibits superior performance across nearly all evaluated setups compared to static $N{:}M$ sparse training (SSN), with the exception of Humanoid-v5 at 1:8 sparsity ("*" in the table). Our method consistently achieves higher average returns with lower variance, confirming that dynamic mask updates significantly improve upon static initialization.  Notably, at 2:4 sparsity, RNM-TD3 outperforms the dense baseline in all environments and remains competitive at 1:4 sparsity. 

In most configurations, our structured sparse approach yields performance gains over unstructured dynamic methods (DS-TD3 and RLx2), achieving up to a 25\% improvement in the complex Ant environment at 50\% sparsity. While unstructured methods occasionally perform better in lower-dimensional environments or at extreme sparsity levels, they lack the hardware-acceleration potential inherent to our $N{:}M$ structured approach.

We note that the original RLx2 implementation \cite{tan2022rlx2} reports higher scores by utilizing a dynamic replay buffer and 3-step returns after 300K steps. However, to isolate the effect of the sparsity mechanism itself, we evaluate all methods under a unified, standard TD3 configuration.

\subsection*{Ablation Study: Impact of Initialization on N:M Sparse Training}

To validate the effectiveness of the proposed initialization strategy, we compare our method, which adjusts the Kaiming distribution to account for the effective fan-in of $N{:}M$ sparsity patterns, against initialization sampling from the standard Kaiming distribution used for dense networks. Adjusting for effective fan-in results in weight values that are scaled by a factor of $\sqrt{M/N}$ at initialization. The results, normalized to the dense baseline, are presented in Table~\ref{tab:Kaiming}.

The most significant observation across environments and sparsity levels is the reduction in variance when using the $N{:}M$ adjusted initialization. However, under high sparsity regimes and in unstable environments (e.g., \textit{Humanoid}), $\sqrt{M/N}$ scaling can produce initial weights with excessively large magnitudes, causing the actor’s outputs to saturate shortly after training begins. This saturation leads to the vanishing gradients problem. When neurons operate in the saturated regime of the tanh nonlinearity, where the derivative approaches zero, gradients diminish significantly during backpropagation, effectively preventing weight updates in the earlier layers.

\begin{table}[t]
\caption{Performance comparison between Kaiming initialization adjusted for N:M fan-in (N:M Adj.) against standard Kaiming initialization (Standard K.). All results are reported as normalized mean scores $\pm$ normalized standard deviation relative to the dense baselines. A mean $>1.0$ indicates performance superior to the dense model.}
\label{tab:Kaiming}
\centering
\renewcommand{\arraystretch}{1}
{\fontsize{9}{11}\selectfont
\begin{tabular}{|c|c|c|c|}
\hline
Env. & Spar. & N:M Adj. & Standard K. \\ \thickhline
\multirow{3}{*}{Ant} & (2:4) & \(\,1.14\,\pm\,0.05\,\) & \(\,1.09\,\pm\,0.45\,\) \\ \cline{2-4}
                     & (1:4) & \(\,1.07\,\pm\,0.15\,\) & \(\,0.94\,\pm\,0.39\,\) \\ \cline{2-4}
                     & (1:8) & \(\,0.86\,\pm\,0.16\,\) & \(\,0.75\,\pm\,0.44\,\) \\ \thickhline
\multirow{3}{*}{HC}  & (2:4) & \(\,1.01\,\pm\,0.08\,\) & \(\,0.96\,\pm\,1.01\,\) \\ \cline{2-4}
                     & (1:4) & \(\,0.90\,\pm\,0.05\,\) & \(\,0.88\,\pm\,0.50\,\) \\ \cline{2-4}
                     & (1:8) & \(\,0.72\,\pm\,0.06\,\) & \(\,0.71\,\pm\,0.61\,\) \\ \thickhline
\multirow{3}{*}{Hum.}& (2:4) & \(\,1.05\,\pm\,0.03\,\) & \(\,1.04\,\pm\,0.37\,\) \\ \cline{2-4}
                     & (1:4) & \(\,1.04\,\pm\,0.03\,\) & \(\,1.03\,\pm\,0.15\,\) \\ \cline{2-4}
                     & (1:8) & \(\,0.28\,\pm\,0.20\,\) & \(\,0.90\,\pm\,1.10\,\) \\ \thickhline
\end{tabular}}
\end{table}

\begin{figure*}[ht]
\vskip 0.2in
\begin{center}
\centerline{\includegraphics[width=\linewidth]{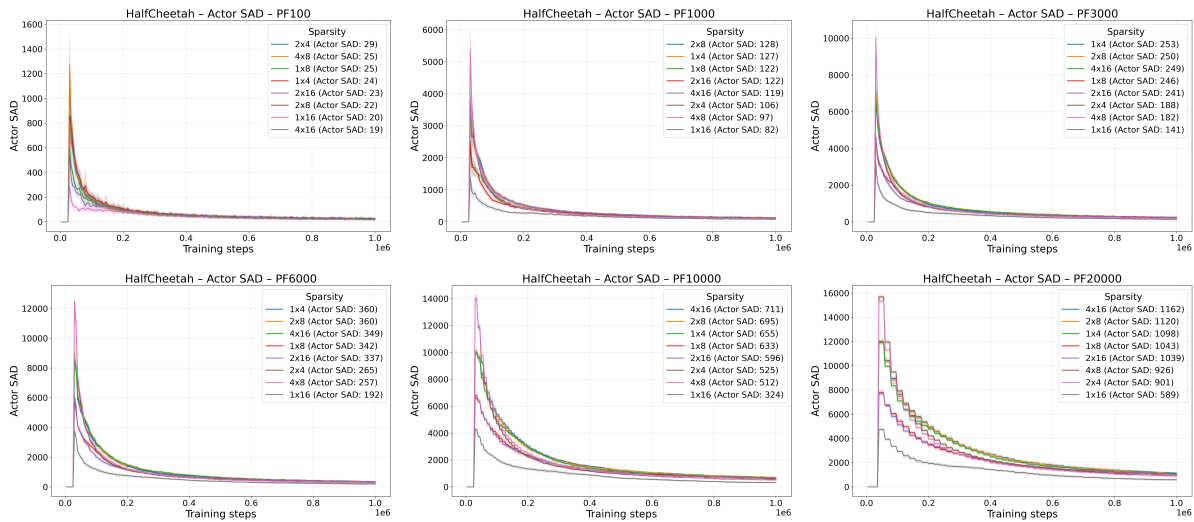}}
\caption{SAD for the HalfCheetah actor network across different N{:}M sparsity patterns and mask update periods. Each curve is averaged over 11 seeds. Across 462 runs, the sparsity level has only a minor effect on SAD for a fixed mask-update period}
\label{fig:HC PF correlation}
\end{center}
\vskip -0.2in
\end{figure*}

\subsection{Ablation study on mask update period}
\label{SAD}
We investigate the relationship between the mask update period, SAD, and agent performance. We vary the mask update period from 10 to 20,000 environment steps in the HalfCheetah environment using a 1:4 sparsity pattern. Results are averaged over 21 random seeds for 1M steps.

\begin{table}[ht]
\caption{Impact of the mask update period on HalfCheetah performance and SAD with 1:4 sparsity.}\label{tab:sparsity_halfcheetah}
\centering
{\fontsize{9}{11}\selectfont
\begin{tabular}{|c|c|c|c|}
  \hline
MA period & Avg. $\pm$ Std. & Act. SAD & Crit. SAD \\ \hline
10        & 7895 $\pm$ 816  & 6        & 6         \\ \hline
50        & 8251 $\pm$ 570  & 17       & 21        \\ \hline
100       & 8298 $\pm$ 895  & 26       & 36        \\ \hline
1000      & 8057 $\pm$ 1120 & 128      & 185       \\ \hline
2000      & 8286 $\pm$ 1139 & 216      & 307       \\ \hline
4000      & 9036 $\pm$ 534  & 346      & 495       \\ \hline
6000      & 8861 $\pm$ 805  & 477      & 679       \\ \hline
8000      & 8717 $\pm$ 753  & 566      & 809       \\ \hline
10000     & 8453 $\pm$ 782  & 654      & 935       \\ \hline
15000     & 8568 $\pm$ 817  & 867      & 1243      \\ \hline
20000     & 8570 $\pm$ 634  & 1071     & 1469      \\ \hline
\end{tabular}
}
\end{table}

Table \ref{tab:sparsity_halfcheetah} reveals a clear trend: as the mask update period increases, the average return improves, peaking at 4{,}000 environment steps before degrading. Crucially, the worst-performing configurations, which correspond to very frequent updates, are associated with very low SAD values. The intuition behind the observed performance trend is as follows:
\begin{itemize}
    \item \textbf{Too small update period:} the network architecture changes too frequently, and the mask is updated before the underlying weights can sufficiently adapt.
    \item \textbf{Too large update period:} the architecture adapts too slowly, and each mask update produces disruptive jumps in average return.
    \item \textbf{Intermediate update period:} strikes a balance between stability and adaptability, allowing the sparse topology to remain stable long enough to learn while still updating frequently enough to remain responsive, resulting in peak performance.
\end{itemize}

An additional pattern can be observed: SAD appears to be correlated with the mask update period. To test this hypothesis, we conducted experiments in all four environments across eight different sparsity patterns. Figure \ref{fig:HC Actor SAD} shows the SAD during the training for the actor network in the HalfCheetah environment. We observe that low SAD correlates with high frequency mask update periods. The same qualitative behavior is observed across all environments and for both actor and critic networks. Furthermore, Figure \ref{fig:HC PF correlation} indicates that, for a fixed mask update period, the sparsity level has only a minor effect on SAD. This implies that the choice of mask update period has more influence on SAD than the sparsity level.

\textbf{Computational Efficiency Analysis.} Although our implementation does not utilize bidirectional $N{:}M$ sparse masks (i.e., row-wise and column-wise N:M sparsity pattern), the computational cost is dominated by forward passes. In our setup (one gradient update per environment step and a policy delay of two), each environment step induces 7 forward passes and 2.5 backward passes across the actor, critics, and target networks. Consequently, even if hardware acceleration is applied only to the forward pass, the end-to-end training speedup remains substantial, but lower than the theoretical $M/N$ factor.

\section{\uppercase{Conclusions \& Future Work}}
\label{sec:conclusion}

In this work, we introduce RNM-TD3, a semi-structured sparse DRL algorithm that maintains strict $N{:}M$ sparsity throughout training. By enforcing a hardware-aware sparsity pattern across the actor and critic networks, our method is aligned with the sparse matrix-matrix multiplication accelerators. Given that the computational cost of TD3 is dominated by forward passes, our approach offers practical end-to-end training speedups, addressing a key limitation of unstructured sparse methods, which often fail to translate theoretical FLOP reductions into real-world acceleration.

Our empirical analysis highlights the critical role of mask update dynamics. We found that a moderate, non-zero Sparse Architecture Divergence (SAD) correlates strongly with improved performance. We hypothesize that maintaining a higher SAD helps preserve topological flexibility, which may be beneficial in long-term training regimes where plasticity typically diminishes. Furthermore, in environments with changing dynamics, agents with larger SAD are expected to exhibit greater adaptability.

Finally, we observe that in high sparsity regimes, weight initialization plays a crucial role, making the early training phase particularly fragile. To address this, future work should investigate novel initialization schemes specifically tailored for high-sparsity regimes in complex environments and adaptive mask update schedules.

\section*{\uppercase{Acknowledgements}}
Parts of this work were supported by the project EFBL, funded by the Bavarian State Ministry for Science and the Arts, and by the High-Tech Agenda of Bavaria.

\bibliographystyle{apalike}
{\small
\bibliography{example}}

\end{document}